%% file: tio-shacl-paper.tex
\documentclass[runningheads]{llncs}

\usepackage[T1]{fontenc}
\usepackage{lmodern}

\usepackage{microtype}

\usepackage{graphicx}
\usepackage{subcaption}

\usepackage{booktabs}
\usepackage{tabularx}
\usepackage{multirow}

\usepackage{listings}
\usepackage{listings-owl}
\usepackage{listings-turtle-paper}
\usepackage{listings-sparql-paper}

\usepackage{hyperref}
\hypersetup{colorlinks=true,linkcolor=black,citecolor=black,urlcolor=blue}
\usepackage{url}

\usepackage{amsmath}
\usepackage{amssymb}

\usepackage{xcolor}
\definecolor{hlremove}{HTML}{FDE8E8}

\graphicspath{{./}}

\newcommand{\tioshacl}{\textsc{tio-shacl}}
\newcommand{\tio}{\textsc{tio}}
\newcommand{\shacl}{\textsc{shacl}}
\newcommand{\shaclaf}{\textsc{shacl-af}}
\newcommand{\sparql}{\textsc{sparql}}
\newcommand{\rdf}{\textsc{rdf}}
\newcommand{\rdfs}{\textsc{rdfs}}
\newcommand{\owl}{\textsc{owl}}
\newcommand{\tmf}{\textsc{tmf}}
\newcommand{\wc}{\textsc{w3c}}
\newcommand{\ibn}{\textsc{ibn}}
\newcommand{\jvm}{\textsc{jvm}}

\newcommand{\MIT}{\textsc{mit}}

\begin{document}

\title{TIO-SHACL: Comprehensive SHACL validation for TMF Intent Ontologies}

\titlerunning{TIO-SHACL: Validation for TMF Intent Ontology}

\author{Jean Martins \and Leonid Mokrushin \and Marin Orlic}
\authorrunning{J. Martins et al.}
\institute{Ericsson Research \\ \email{\{jean.martins, leonid.mokrushin, marin.orlic\}@ericsson.com}}

\maketitle

\input{content}

\section*{Declaration of Use of Generative AI}

Generative AI tools were used during the development of this work. Specifically: (1)~the \tioshacl{} validation framework (SHACL shapes, SPARQL constraints, test cases, and Python tooling) was co-developed with AI coding assistants; (2)~AI tools assisted with drafting, restructuring, and editing paper text; (3)~benchmark experiment scripts and plot generation code were co-developed with AI assistance. All AI-generated content was reviewed, validated, and revised by the authors, who take full responsibility for the correctness and originality of the work.

\input{tio-shacl-paper.bbl}
\end{document}

%% file: content.tex
\begin{abstract}
Intent-based networking promises to revolutionize telecommunications network management by enabling operators to specify high-level goals rather than low-level configurations. The TM Forum Intent Ontology (\tio{}) provides standardized vocabulary for expressing network intents, yet lacks formal validation mechanisms to ensure intent correctness before its admission. We present \tioshacl{}, the first comprehensive \shacl{} (Shapes Constraint Language) validation framework for the \tmf{} Intent Ontology. Our contribution includes 56 node shapes and 69 property shapes across all 15 \tio{} v3.6.0 ontology modules, a reusable constraint library with 25 parameterized \sparql{}-based constraint components, and novel validation patterns for recursive logical operators, quantity-based constraints, and cross-expectation relationships. We pursued 100\% vocabulary coverage (87 classes, 109 properties, 72 functions), cross-implementation compatibility across three major \shacl{} engines, and validation accuracy on a corpus of 133 test cases. \tioshacl{} is publicly available under \MIT{} license at \url{https://github.com/EricssonResearch/tio-shacl} and enables automated syntactic and semantic validation of network intents, addressing a critical gap in the field.

\end{abstract}

\section{Introduction}
\label{sec:introduction}

The telecommunications industry is undergoing a fundamental transformation toward autonomous networks capable of self-configuration, self-optimization, and self-healing~\cite{tmf-an}. At the heart of this transformation lies \emph{intent-based networking} (\ibn{}), a paradigm that allows network operators to specify desired outcomes, such as ``ensure 99.99\% availability for video streaming services'', rather than explicit device configurations~\cite{rfc9315}. The TM Forum (\tmf{}) work body on intent-based automation has developed a comprehensive Intent Ontology (\tio{}) to standardize how intents are expressed and reported~\cite{tmf-tr290}.

Despite the maturity of the \tio{} specification, spanning 15 ontology modules with 87 classes and 109 properties, a critical gap remains: no formal validation mechanism exists for ensuring intent correctness. Without standardized validation, syntactically valid \rdf{} that violates \tio{} semantics may be silently accepted by intent handlers, and intent portability across vendors cannot be guaranteed. Moreover, the inherent complexity of \tio{}'s recursive logical operators, quantity expressions, and validity constraints creates a steep learning curve that hampers its faster adoption.

We address this gap by presenting \tioshacl{}, the first comprehensive \shacl{} validation framework for the \tmf{} Intent Ontology. \shacl{} (Shapes Constraint Language)~\cite{w3c-shacl} is a \wc{} standard for validating \rdf{} graphs against structural and semantic constraints. Our work makes five key contributions. First, we provide a \emph{complete \shacl{} formalization} of \tio{} v3.6.0, covering all 15 ontology modules with 56 node shapes, 69 property shapes, and 147 constraint instances. Second, we develop a \emph{reusable constraint library} with 25 parameterized \sparql{}-based constraint components and 3 custom target types, enabling modular and maintainable validation. Third, we introduce \emph{novel validation patterns} for \tio{}-specific constructs including recursive logical operators (\texttt{log:allOf}, \texttt{log:anyOf}), quantity comparisons with units, and cross-expectation validity inheritance. Fourth, we present \emph{empirical validation} demonstrating vocabulary coverage, cross-implementation compatibility across three major \shacl{} engines, and validation accuracy on 133 test cases. Finally, we provide an \emph{open-source release}\footnote{\url{https://github.com/EricssonResearch/tio-shacl}} of all shapes, test cases, and tooling to enable community adoption and reproducibility.

A distinguishing aspect of our approach is its \emph{test-driven development methodology}. The 133-case test corpus, balanced between valid and invalid intents, serves not only as a correctness benchmark but also as a regression safety net for ontology evolution. When a new \tio{} version introduces classes, properties, or revised semantics, the existing test suite immediately surfaces shapes that require updating, while confirming that unaffected modules remain correct. This test-driven workflow substantially lowers the cost of adapting \tioshacl{} to future \tio{} releases: developers extend or modify shapes incrementally, guided by failing tests rather than manual specification diffing, and can verify full backward compatibility before deployment.

The remainder of this paper is organized as follows: Section~\ref{sec:background} provides background on \tio{} and \shacl{}. Section~\ref{sec:methodology} describes our shape development methodology. Section~\ref{sec:shapes} details the shape architecture and novel patterns. Section~\ref{sec:evaluation} presents our empirical evaluation, and Section~\ref{sec:conclusion} concludes.

\section{Background}
\label{sec:background}

\subsection{TMF Intent Ontology (TIO)}

The TM Forum Intent Ontology~\cite{tmf-tr292} provides standardized vocabulary for expressing network management intents. \tio{} v3.6.0 is specified across a series of Technical Reports: TR290~\cite{tmf-tr290} defines the Intent Common Model, TR292A--E~\cite{tmf-tr292a,tmf-tr292b,tmf-tr292c,tmf-tr292d,tmf-tr292e} cover management elements, state machines, function definitions, quantity ontology, and logical operators, and the TR291 series~\cite{tmf-tr291a,tmf-tr291c,tmf-tr291g,tmf-tr291h,tmf-tr291i} adds extension modules for validity, guarantees, probing, preferences, and utility. TR299~\cite{tmf-tr299} defines intent specifications, while IG1253~\cite{tmf-ig1253} and IG1358~\cite{tmf-ig1358} provide usage patterns and operational guidance. Together, these documents define 15 ontology modules delivered as normative \rdf{} (Turtle) files. \tio{} follows a layered architecture. Figure~\ref{fig:intent-example} shows a typical \tio{} intent requesting a generic video service with an specializing throughput constraint.

\begin{figure}[t]
\begin{lstlisting}[language=TurtlePaper,caption={TIO intent for video service with throughput constraint},label={fig:intent-example}]
@prefix icm:  <http://tio.models.../v3.6.0/IntentCommonModel/> .
@prefix log:  <http://tio.models.../v3.6.0/LogicalOperators/> .
@prefix met:  <http://tio.models.../v3.6.0/MetricsAndObservations/> .
@prefix quan: <http://tio.models.../v3.6.0/QuantityOntology/> .
@prefix set:  <http://tio.models.../v3.6.0/SetOperators/> .

ex:VideoIntent a icm:Intent ;
    log:allOf ( ex:DeliveryExp ex:ThroughputExp ) .

ex:DeliveryExp a icm:DeliveryExpectation ;
    icm:target ex:VideoTarget ;
    icm:deliveryType cfss:VideoCFSS .

ex:VideoTarget a icm:Target ;
    icm:chooseFrom [ set:resourcesOfType ( cfss:VideoCFSS ) ] .

ex:ThroughputCond a log:Condition ;
    quan:atLeast ( [ met:lastValue ( dim:Throughput ) ] "320kbps"^^quan:quantity ) .

ex:ThroughputExp a icm:PropertyExpectation ;
    icm:target ex:VideoTarget ;
    log:allOf ( ex:ThroughputCond ) .
\end{lstlisting}
\end{figure}

Intent Common Model defines core constructs through classes such as \texttt{icm:Intent} for top-level containers, \texttt{icm:Expectation} for requirements, \texttt{icm:Target} for scope, and \texttt{icm:Report} for feedback. Expectations are specialized into \texttt{DeliveryExpectation}, \texttt{PropertyExpectation}, and \texttt{ReportingExpectation}, respectively covering service provisioning, property constraints, and monitoring requirements.

Logical Operators define compositional requirements through classes such as \texttt{log:Condition} for atomic constraints, and functions such as \texttt{log:allOf}, \texttt{log:anyOf}, and \texttt{log:match} for conjunction, disjunction, and exact value matching, respectively. Together, these enable intents to combine multiple requirements into complex boolean expressions.

Quantity Ontology defines metric-based constraints through classes such as \texttt{quan:Quantity} for unit-aware values, and functions such as \texttt{quan:atLeast}, \texttt{quan:atMost}, \texttt{quan:exactly}, and \texttt{quan:between}, respectively enabling comparison of observed metrics against declared thresholds with unit handling.\footnote{The TR291 series~\cite{tmf-tr291a,tmf-tr291c,tmf-tr291g,tmf-tr291h,tmf-tr291i} adds extension modules for validity constraints, intent negotiation, probing, and handling preferences.}

\subsection{SHACL: Shapes Constraint Language}

\shacl{}~\cite{w3c-shacl} is a \wc{} recommendation for describing and validating \rdf{} graphs. The language centers on two primary constructs: \emph{node shapes} (\texttt{sh:NodeShape}) that define constraints on nodes of specific types or matching specific patterns, and \emph{property shapes} (\texttt{sh:PropertyShape}) that constrain the values of specific properties. \shacl{} provides a rich vocabulary of \emph{constraint components} including cardinality restrictions (\texttt{sh:minCount}, \texttt{sh:maxCount}), value type requirements (\texttt{sh:class}, \texttt{sh:datatype}), enumeration membership (\texttt{sh:in}), and string pattern matching (\texttt{sh:pattern}). Validation produces a conformance report indicating whether a data graph satisfies all applicable shapes, with detailed violation messages that facilitate debugging and correction.

\shacl{} Advanced Features (\shaclaf{}) go beyond core \shacl{} by defining three extension mechanisms that \tioshacl{} relies upon. \emph{\sparql{}-based constraints} (\texttt{sh:sparql}) embed arbitrary \sparql{} queries within shapes for validation logic such as function arity checking and cross-reference resolution. \emph{\sparql{}-based target types} (\texttt{sh:SPARQLTargetType}) parameterize node selection queries for reuse across shapes, avoiding duplication of complex targeting logic. \emph{Constraint components} (\texttt{sh:ConstraintComponent}) encapsulate validation logic into reusable, parameterized validators that can be instantiated with different parameters across shapes. Together, these mechanisms enable parameterized, reusable validation logic for \tio{}'s 72+ functions and 15 ontology modules. Notably, \shaclaf{} support varies across implementations: Apache Jena is limited to the \sparql{} tier and does not support custom target types or constraint components, a practical consideration we revisit in our cross-implementation evaluation (Section~\ref{sec:evaluation}).

\subsection{Validation by Example}
\label{sec:validation-example}

To illustrate how \shacl{} validation works in practice for \tio{} intents, we walk through a concrete example involving quantity comparison functions, one of \tio{}'s most common patterns. Listing~\ref{lst:func-defs} shows how they are defined.
\begin{figure}[tp]
\begin{lstlisting}[language=TurtlePaper,caption={\tio{} function definitions for quantity comparison (abbreviated)},label={lst:func-defs}]
quan:atLeast a fun:Function ;
  fun:resultType xsd:boolean ;
  fun:argumentTypes ( quan:Quantity quan:Quantity ) .

met:lastValue a fun:Function ;
  fun:resultType rdf:Resource ;  # polymorphic
  fun:argumentTypes ( met:Metric ) .
\end{lstlisting}
\end{figure}

Function Definitions in the Ontology. \tio{} functions as first-class entities with declared arity and type signatures. Above we see \texttt{quan:atLeast} which compares two quantities and returns a boolean, while \texttt{met:lastValue} retrieves the most recent observation for a metric. Notably, \texttt{met:lastValue} declares \texttt{fun:resultType rdf:Resource}, a polymorphic return type whose actual type is inferred from the metric's \texttt{rdfs:range} at validation time.

Correct Usage: Listing~\ref{lst:valid-usage} shows the idiomatic pattern: \texttt{dim:Throughput} is declared as a \texttt{met:Metric} with \texttt{rdfs:range quan:Quantity}, so \texttt{met:lastValue} accepts it and its result type flows through as a \texttt{quan:Quantity} to the outer \texttt{quan:atLeast} comparison. The quantity shorthand literal \texttt{"320kbps"\^{}\^{}quan:quantity} serves as the threshold.
\begin{lstlisting}[language=TurtlePaper,caption={Removing the highlighted declarations triggers violations.},label={lst:valid-usage},escapechar=!]
dim:Throughput !\colorbox{hlremove}{\texttt{a met:Metric ;}}!
    !\colorbox{hlremove}{\texttt{rdfs:range quan:Quantity }}!.
ex:ThroughputCond
    quan:atLeast ( [ met:lastValue ( dim:Throughput ) ] 
                    "320kbps"^^quan:quantity ) .
\end{lstlisting}

A Subtle Error: If the highlighted declarations \texttt{a met:Metric} and \texttt{rdfs:range quan:Quantity} are omitted, the \rdf{} remains syntactically valid, \texttt{dim:Throughput} is simply an untyped resource, but \texttt{met:lastValue} expects a \texttt{met:Metric} argument (Listing~\ref{lst:func-defs}). Without the type declaration, the validator cannot verify type compatibility or infer the result type for the outer \texttt{quan:atLeast} comparison.

\shacl{} Catches the Error: Running \tioshacl{} validation on the invalid version produces the report in Listing~\ref{lst:shacl-report}. Two violations cascade from the single missing declaration: first, \texttt{met:lastValue} rejects \texttt{dim:Throughput} because it is not a \texttt{met:Metric}; second, \texttt{quan:atLeast} cannot verify its first argument is a \texttt{quan:Quantity}, since the broken type chain prevents result type inference.

\begin{lstlisting}[language=TurtlePaper,caption={SHACL validation report (abbreviated)},label={lst:shacl-report}]
[] a sh:ValidationReport ;
  sh:conforms false ;
  sh:result [
    sh:focusNode _:lastValueCall ;
    sh:resultSeverity sh:Violation ;
    sh:sourceConstraint  tio:FunctionUsageArgumentTypeObjectConstraint ;
    sh:resultMessage "Function met:lastValue expects met:Metric." ] ;
  sh:result [
    sh:focusNode ex:ThroughputCond ;
    sh:resultSeverity sh:Violation ;
    sh:sourceConstraint tio:FunctionUsageArgumentTypeObjectConstraint ;
    sh:resultMessage "Function quan:atLeast expects quan:Quantity." ] .
\end{lstlisting}

This example illustrates how structurally valid \rdf{} can be semantically incorrect, and how \shacl{} validation catches such errors with actionable messages. The remainder of this paper describes how \tioshacl{} scales this approach across all \tio{} ontology modules, classes, and functions.

\section{Shape Development Methodology}
\label{sec:methodology}

\subsection{Design Principles}

We developed \tioshacl{} following four guiding principles. \emph{Completeness} ensures that all \tio{} classes and properties receive appropriate validation constraints, leaving no semantic gaps that could permit invalid intents. \emph{Modularity} enables constraint reuse through parameterized validators and custom target types, reducing duplication and simplifying maintenance as \tio{} evolves. \emph{Standards compliance} restricts our implementation to \shacl{}-Core and \shacl{}-\sparql{} features supported by major implementations, ensuring broad tool compatibility without vendor lock-in. Finally, \emph{actionability} requires that violation messages provide sufficient context to guide correction, including the specific constraint violated and the offending value.

\subsection{Development Process}

Our development followed an iterative process spanning five phases. We began with \emph{ontology analysis}, systematically extracting classes, properties, domain/range declarations, and cardinality restrictions from the normative Turtle files across all 15 \tio{} modules. This was followed by \emph{specification mining}, where we analyzed \tmf{} Technical Reports (TR290-TR294 series) to identify implicit constraints not formally encoded in the ontology, for example, the requirement that Conditions must appear within Expectations rather than directly on Intents.

During \emph{shape authoring}, we created shapes following a consistent structural pattern: one NodeShape per \tio{} class with embedded PropertyShapes for class-specific properties. As patterns emerged across modules, we performed \emph{constraint extraction}, factoring common validation logic into a reusable library of parameterized \sparql{} queries. Finally, \emph{test-driven refinement} validated shapes against a growing corpus of both valid and intentionally invalid intent examples, with constraint logic iteratively refined based on false positives and false negatives.

\subsection{Key Design Decisions}

Three design decisions shaped the \tioshacl{} architecture:

\paragraph{Mixin Extensions vs.\ Modifying TIO Baseline}
Certain validation patterns require semantic classifications absent from \tio{} v3.6.0, for example, distinguishing entities that \emph{must} carry boolean functions from those that \emph{may} participate in boolean logic. Rather than modifying \tio{} ontology files directly, we created separate extension files using \texttt{rdfs:subClassOf} mixins. This keeps the \tio{} baseline pristine, enabling clean merging of future \tio{} versions without conflicts, and ensures existing valid intents remain valid since extensions are purely additive.

\paragraph{Parameterized Constraints vs.\ Per-Function Shapes}
\tio{} defines 72+ functions, each requiring arity validation via \rdf{} list element counting. Instead of writing a separate \sparql{} query per function, we developed parameterized \texttt{sh:ConstraintComponent} definitions instantiated per function. A single \texttt{FunctionUsageArityConstraint} validates all functions: when arity logic needs improvement, one change benefits all functions, and new \tio{} functions only require instantiating the existing component with new parameters.

\paragraph{Positive Validation vs.\ Negative Checks}
The Intent\(\rightarrow\)Expectation\(\rightarrow\)Condition hierarchy requires validating that intents reference only expectations (not conditions) in their boolean functions. Rather than enumerating forbidden types (``operand is NOT a Condition''), we introduced an \texttt{icm:IntentOperand} mixin where allowed types opt in. This positive-validation approach is robust to ontology evolution: new expectation types automatically participate if marked as \texttt{IntentOperand}, following the open--closed principle.

\subsection{Extension Vocabulary}
\label{sec:extensions}

During development, we identified semantic gaps in the base \tio{} ontology that hindered complete validation. To address these without modifying the baseline, we introduced mixin extension classes (via \texttt{rdfs:subClassOf}) in separate files. We illustrate two extensions through concrete examples, then summarize the remaining three.

\paragraph{Actionable Validation.}
\tio{} allows structurally valid intents that cannot be operationalized because they lack the boolean functions needed for evaluation. Three mixin classes address this: \texttt{fun:BooleanFunction} marks functions with \texttt{resultType xsd:boolean}; \texttt{fun:Evaluable} marks entities that \emph{can} participate in boolean logic; and \texttt{fun:Actionable} (subclass of \texttt{Evaluable}) marks entities that \emph{must} carry boolean computation. With these classifications, a single \sparql{} constraint validates all actionable entities. Listing~\ref{lst:actionable-example} shows an Intent that is syntactically valid but not executable, removing the highlighted \texttt{log:allOf} line leaves no boolean computation.
\begin{lstlisting}[language=TurtlePaper,caption={Invalid Intent: removing the highlighted line triggers a violation},label={lst:actionable-example},escapechar=!]
ex:myIntent a icm:Intent ;
  rdfs:label "Bandwidth Intent"@en ;
  !\colorbox{hlremove}{\texttt{log:allOf ( ex:E1 ex:E2 ) }}!.  # REQUIRED
\end{lstlisting}

Running \tioshacl{} on the version without \texttt{log:allOf} produces:
\begin{lstlisting}[language=TurtlePaper,caption={Validation report for missing boolean function},label={lst:actionable-report}]
[] a sh:ValidationReport ;
  sh:conforms false ;
  sh:result [
    sh:focusNode ex:myIntent ;
    sh:resultSeverity sh:Violation ;
    sh:sourceConstraint tio:ActionableBooleanEvaluableConstraint ;
    sh:resultMessage "Actionable instance of class icm:Intent missing BooleanFunction property. Add log:allOf, log:anyOf, etc." ] .
\end{lstlisting}

\paragraph{Intent Hierarchy Validation.}
\tio{} enforces a hierarchy: Intents reference Expectations, which in turn reference Conditions. Rather than checking ``operand is NOT a Condition'' (brittle, every new type requires updating the exclusion list), we introduced \texttt{icm:IntentOperand} and \texttt{icm:ExpectationOperand} mixins where allowed types opt in. Listing~\ref{lst:operand-example} shows a common mistake: removing the highlighted Expectation wrapper and placing the Condition directly in the Intent's \texttt{log:allOf}.
\begin{lstlisting}[language=TurtlePaper,caption={Hierarchy violation: removing the highlighted Expectation wrapper triggers a violation},label={lst:operand-example},escapechar=!]
ex:myIntent a icm:Intent ;
  log:allOf ( !\colorbox{hlremove}{\texttt{ex:myExpectation}}! ) .  # CORRECT
  # log:allOf ( ex:myCondition ) .  # WRONG
!\colorbox{hlremove}{\texttt{ex:myExpectation a icm:PropertyExpectation ;}}!
  !\colorbox{hlremove}{\texttt{icm:target ex:T1 ;}}!
  !\colorbox{hlremove}{\texttt{log:allOf ( ex:myCondition ) .}}!
ex:myCondition a log:Condition ;
  log:allOf ( ex:C1 ) .
\end{lstlisting}

Without the Expectation wrapper, \texttt{log:Condition} is not an \texttt{icm:\allowbreak{}Intent\-Operand} and validation fails:
\begin{lstlisting}[language=TurtlePaper,caption={Validation report for hierarchy violation},label={lst:operand-report}]
[] a sh:ValidationReport ;
  sh:conforms false ;
  sh:result [
    sh:focusNode ex:myIntent ;
    sh:resultSeverity sh:Violation ;
    sh:resultMessage "Intent ex:myIntent references non-IntentOperand in log:allOf. Wrap Conditions in PropertyExpectation." ] .
\end{lstlisting}

This positive-validation approach follows the open--closed principle: new expectation types automatically participate if marked as \texttt{IntentOperand}, with no changes to existing constraints.

\paragraph{Remaining Extensions.}
Three additional extensions complete the vocabulary. \texttt{fun:ContainerTyped} marks entities that are or produce \texttt{rdfs:Container} values, replacing complex \sparql{} filters with a single \texttt{sh:class} constraint. \texttt{iv:ValidityCandidate} identifies entities that can have validity assigned (intents and expectations), enabling a simple range constraint for \texttt{iv:sameValidityAs}. Together, these five extension categories enable validation of \emph{operational semantics}, ensuring intents are not merely well-formed \rdf{} but actually executable by intent handlers.

\section{Shape Architecture}
\label{sec:shapes}

\subsection{Overall Structure}

\tioshacl{} comprises 56 node shapes and 69 property shapes across 15 ontology modules, backed by 25 parameterized \sparql{} constraint components and validated against 133 test cases (67 valid, 66 invalid). Table~\ref{tab:shape-stats} summarizes the per-module distribution:

\begin{table}[htbp]
\caption{\tioshacl{} Shape Statistics by Ontology Module}
\label{tab:shape-stats}
\centering
\small
\begin{tabular}{lrrr}
\toprule
\textbf{Ontology Module} & \textbf{Node} & \textbf{Property} & \textbf{SPARQL} \\
\midrule
IntentCommonModel & 12 & 13 & 5 \\
IntentManagementOntology & 7 & 5 & 2 \\
QuantityOntology & 6 & 5 & 2 \\
FunctionOntology & 5 & 2 & 2 \\
PreferenceOfHandlingOutcomes & 4 & 13 & 0 \\
IntentValidityOntology & 4 & 3 & 1 \\
IntentSpecification & 3 & 10 & 0 \\
IntentGuaranteeOntology & 3 & 3 & 0 \\
Utility & 3 & 6 & 0 \\
MathFunctions & 2 & 2 & 0 \\
ProposalBestIntent & 2 & 3 & 0 \\
IntentProbing & 2 & 2 & 0 \\
LogicalOperators & 1 & 0 & 1 \\
MetricsAndObservations & 1 & 2 & 0 \\
SetOperators & 1 & 0 & 1 \\
\midrule
\textbf{Total} & \textbf{56} & \textbf{69} & \textbf{14} \\
\bottomrule
\end{tabular}
\end{table}

Table~\ref{tab:constraint-types} breaks down the 147 constraint instances by \shacl{} constraint type. Type constraints (\texttt{sh:class}: 40, \texttt{sh:datatype}: 23) and structural constraints (\texttt{sh:nodeKind}: 20) dominate, reflecting \tio{}'s emphasis on typed, well-structured \rdf{}. Cardinality constraints (\texttt{sh:maxCount}: 19, \texttt{sh:minCount}: 14) enforce property multiplicities, while \texttt{sh:sparql} (18) handles complex validation logic, such as function arity checking and cross-reference resolution, that exceeds core \shacl{} expressiveness.

\begin{table}[htbp]
\caption{\shacl{} constraint type frequency across all shape files}
\label{tab:constraint-types}
\centering
\small
\begin{tabular}{lr@{\hskip 2em}lr}
\toprule
\textbf{Constraint Type} & \textbf{Count} & \textbf{Constraint Type} & \textbf{Count} \\
\midrule
\texttt{sh:class}        & 40 & \texttt{sh:or}            & 4 \\
\texttt{sh:datatype}     & 23 & \texttt{sh:hasValue}      & 3 \\
\texttt{sh:nodeKind}     & 20 & \texttt{sh:minInclusive}  & 3 \\
\texttt{sh:maxCount}     & 19 & \texttt{sh:and}           & 2 \\
\texttt{sh:sparql}       & 18 & \texttt{sh:pattern}       & 1 \\
\texttt{sh:minCount}     & 14 & & \\
\midrule
\multicolumn{3}{l}{\textbf{Total}} & \textbf{147} \\
\bottomrule
\end{tabular}
\end{table}

\subsection{Novel Validation Patterns}

Three \sparql{}-based patterns address validation challenges that exceed core \shacl{} expressiveness.

\subsubsection{Parameterized Function Arity}

\tio{} defines 72+ functions, each with declared arity. A single \texttt{sh:SPARQLConstraint} validates all of them, the \sparql{} query in Listing~\ref{lst:arity-constraint} is embedded in the constraint definition and instantiated per function:

\begin{lstlisting}[language=SPARQLPaper,caption={Parameterized arity validation via RDF list traversal},label={lst:arity-constraint}]
SELECT $this ?func ?arityMin ?arityMax ?actualCount
WHERE {
  # Match any function call and its declared arity
  $this ?func ?argList .
  ?func a fun:Function ;
        fun:arityMin ?arityMin .
  OPTIONAL { ?func fun:arityMax ?arityMax }
  # Count elements by traversing the RDF list
  OPTIONAL {
    SELECT ?argList (COUNT(?item) AS ?countFromList)
    WHERE { ?argList rdf:rest*/rdf:first ?item }
    GROUP BY ?argList
  }
  # Empty list (rdf:nil) yields no count; treat as 0
  BIND(IF(?argList = rdf:nil, 0, ?countFromList)
       AS ?actualCount)
  # Report if outside [arityMin, arityMax] bounds
  FILTER(?actualCount < ?arityMin ||
    (BOUND(?arityMax) && ?actualCount > ?arityMax))
}
\end{lstlisting}

One change to this constraint benefits all 72+ functions; new functions only require instantiation with new parameters.

\subsubsection{Recursive Logical Operator Validation}

\tio{} logical operators (\texttt{log:allOf}, \texttt{log:anyOf}) accept \rdf{} lists of arbitrary depth. Each argument must be boolean-evaluable, but the types that qualify are heterogeneous:

\begin{lstlisting}[language=TurtlePaper,caption={Recursive boolean argument validation},label={lst:recursive}]
SELECT $this ?badArg
WHERE {
  # Traverse nested list to reach each argument
  $this log:allOf/rdf:rest*/rdf:first ?arg .
  # Argument must be one of three boolean-evaluable kinds
  FILTER NOT EXISTS {
    { ?arg a fun:Evaluable }       # e.g. Expectation
    UNION
    { ?arg a fun:BooleanFunction } # e.g. quan:atLeast
    UNION
    { FILTER(isLiteral(?arg)       # literal true/false
        && datatype(?arg) = xsd:boolean) }
  }
  BIND(?arg AS ?badArg)
}
\end{lstlisting}

The \texttt{rdf:rest*/rdf:first} path flattens arbitrarily nested lists in a single traversal. The \texttt{fun:Evaluable} mixin (Section~\ref{sec:extensions}) avoids enumerating concrete classes, new evaluable types are covered automatically.

\subsubsection{Vocabulary Spell-Checking}

\rdf{}'s open-world assumption means a typo like \texttt{icm:taget} (instead of \texttt{icm:target}) is valid \rdf{} but semantically wrong, the intent silently lacks a target. No standard \rdf{} or \owl{} mechanism catches this. We define an \texttt{sh:SPARQLConstraint} that flags properties from \tio{} namespaces not declared in the ontology:

\begin{lstlisting}[language=SPARQLPaper,caption={Vocabulary usage constraint (``spell checker'' for TIO namespaces)},label={lst:vocab-constraint}]
SELECT $this ?prop
WHERE {
  $this ?prop ?value .
  # Only check properties in TIO namespaces
  FILTER(STRSTARTS(STR(?prop), $ontologyNamespace))
  # Flag if not declared in the ontology
  FILTER NOT EXISTS { ?prop a rdf:Property . }
}
\end{lstlisting}

\section{Evaluation}
\label{sec:evaluation}

We evaluate \tioshacl{} along four dimensions: vocabulary coverage, test methodology and results, cross-implementation compatibility, and performance.

\subsection{Coverage Analysis}

A coverage report generator systematically analyzes which \tio{} vocabulary elements are exercised by \shacl{} shapes and test cases. \tioshacl{} achieves 100\% coverage of \tio{} v3.6.0: all 87 classes have corresponding NodeShapes, all 109 properties have PropertyShapes or \sparql{} constraints, and all 72 functions are validated for arity, result type, and argument types via the parameterized constraint library.

\subsection{Test Methodology and Results}

\tioshacl{} uses 133 self-contained Turtle test files organized by ontology module, split nearly equally between 67 valid (``good'') and 66 invalid (``bad'') examples. Good tests demonstrate correct usage patterns and must pass with \texttt{sh:conforms~=~true}; bad tests contain intentional violations, missing properties, type mismatches, arity errors, invalid vocabulary, and must produce expected violation reports. The near-equal split ensures both precision (valid intents not rejected) and recall (invalid intents detected). All 133 test cases pass their expected outcomes.

\subsection{Cross-Implementation Compatibility}

\tioshacl{} achieves 100\% pass rate and agreement across three major \shacl{} implementations: pySHACL~0.31.0, TopBraid~\shacl{}~1.4.3, and Apache~Jena~5.2.0. An initial benchmark run revealed 32 disagreements where Apache Jena flagged valid test files as violations. The root cause is Jena's built-in \texttt{VocabularyUsageShape}, which requires explicit \texttt{rdf:type} declarations for resources appearing in function-call positions (e.g., arguments to \texttt{quan:atLeast}); pySHACL and TopBraid tolerate implicit typing in these contexts. After adding explicit \texttt{rdf:type} declarations to the 36 affected test files, all three validators achieve 100\% agreement. Cross-validator agreement was verified on the function-related test suites (FunctionOntology, LogicalOperators, MathFunctions, QuantityOntology, SetOperators, 37 test files, 100\% agreement).

Notable implementation-specific findings:
\begin{itemize}
    \item \textbf{pySHACL}: \rdfs{} inference must be disabled (\texttt{inference=None}) to prevent incorrect domain/range class assignments to mislead the results.
    \item \textbf{TopBraid}: Cached as a singleton to amortize \jvm{} startup cost, yielding more consistent per-file timing.
    \item \textbf{Jena}: Fastest validator overall but limited to the \sparql{} tier (no \shaclaf{} support), requiring inline \texttt{sh:SPARQLTarget} queries.
\end{itemize}

\subsection{Performance Analysis}

We benchmarked all 133 test cases against three \shacl{} engines on the \sparql{} feature tier. Each (file, validator) pair was measured over 6 repetitions preceded by 2 warmup runs. For Java validators (Jena and TopBraid), we use internal timing: file loading and graph parsing occur before timing starts, so only the \shacl{} validation call is measured. Experiments ran on Linux x86/64 (WSL2), 4 logical cores, 23.5\,GB RAM, Python~3.11, OpenJDK~21.

\begin{figure}[tbp]
\centering
\begin{subfigure}[t]{0.48\linewidth}
\centering
\includegraphics[width=\linewidth]{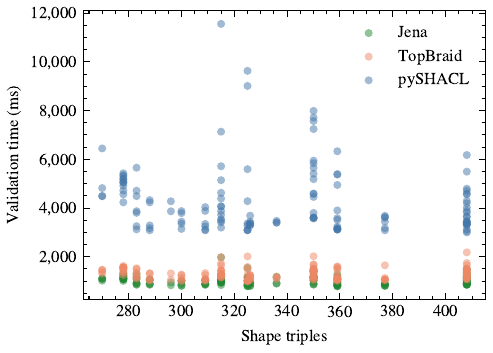}
\caption{Validation time vs.\ number of triples. }
\label{fig:time-vs-triples}
\end{subfigure}
\hfill
\begin{subfigure}[t]{0.48\linewidth}
\centering
\includegraphics[width=\linewidth]{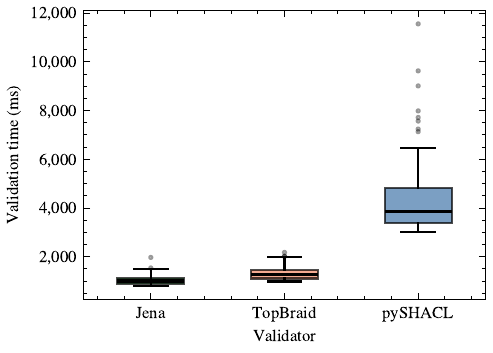}
\caption{Time distribution by backend. }
\label{fig:time-distribution}
\end{subfigure}
\caption{Performance characteristics across three \shacl{} engines.}
\label{fig:perf-plots}
\end{figure}

Table~\ref{tab:perf-summary} summarizes aggregate performance. All pairwise differences are statistically significant (Mann-Whitney~$U$, $p < 0.001$) with large effect sizes (Cohen's~$d > 1.8$). Jena is 4.2$\times$ faster than pySHACL and 1.3$\times$ faster than TopBraid. pySHACL exhibits substantially higher variance (Figure~\ref{fig:time-distribution}), indicating less predictable execution across varying test complexity.

\begin{table}[htbp]
\caption{Validation performance across 133 test cases (\sparql{} tier).}
\label{tab:perf-summary}
\centering
\begin{tabular}{lrrrr}
\toprule
\textbf{Validator} & \textbf{Total (s)} & \textbf{Avg/file (ms)} & \textbf{Std (ms)} & \textbf{Speedup} \\
\midrule
Jena 5.2.0 & 136.1 & 1,024 & 185 & 4.2$\times$ \\
TopBraid \shacl{} 1.4.3 & 174.4 & 1,312 & 235 & 3.3$\times$ \\
pySHACL 0.31.0 & 573.9 & 4,315 & 1,398 & 1.0$\times$ \\
\bottomrule
\end{tabular}
\end{table}

Table~\ref{tab:perf-module} breaks down performance by ontology module for five representative modules. Jena speedups range from 3.4$\times$ (FunctionOntology) to 5.0$\times$ (IntentSpecification). Modules exercising recursive list traversal, i.e., QuantityOntology, LogicalOperators, SetOperators, show elevated pySHACL times (above 5,000\,ms), consistent with higher per-constraint overhead in \sparql{} engine. Figure~\ref{fig:time-vs-triples} confirms that pySHACL's cost grows more steeply with input size.

\begin{table}[htbp]
\caption{Per-module validation performance for 5 representative modules (\sparql{} tier, internal timing). Speedup is Jena relative to pySHACL.}
\label{tab:perf-module}
\centering
\small
\begin{tabular}{lrrrrr}
\toprule
\textbf{Module} & \textbf{Tests} & \textbf{Jena (ms)} & \textbf{TopBraid (ms)} & \textbf{pySHACL (ms)} & \textbf{Speedup} \\
\midrule
QuantityOntology    & 17 & 1,149 & 1,398 & 5,315 & 4.6$\times$ \\
IntentSpecification &  9 &   996 & 1,267 & 4,941 & 5.0$\times$ \\
IntentCommonModel   & 24 & 1,033 & 1,375 & 3,920 & 3.8$\times$ \\
FunctionOntology    &  3 &   996 & 1,143 & 3,425 & 3.4$\times$ \\
Utility             &  6 &   886 & 1,165 & 3,406 & 3.8$\times$ \\
\bottomrule
\end{tabular}
\end{table}

For validators supporting \shacl{} Advanced Features (pySHACL and TopBraid), we also measured the AF tier, which replaces inline \texttt{sh:SPARQLTarget} queries with parameterized \texttt{sh:SPARQLTargetType} declarations. The AF tier adds less than 2\% overhead: pySHACL averages 4,355\,ms/file (+0.9\%) and TopBraid 1,332\,ms/file (+1.5\%) compared to the \sparql{} tier. This confirms that the parameterized target type indirection in \shaclaf{} adds negligible overhead, justifying the use of Advanced Features for maintainability.

\section{Conclusion}
\label{sec:conclusion}

We presented \tioshacl{}, the first comprehensive \shacl{} validation framework for the \tmf{} Intent Ontology. Our contribution of 56 node shapes and 69 property shapes across 15 ontology modules enables automated validation of network intents with 100\% vocabulary coverage, addressing a critical gap in autonomous network adoption. The modular architecture with 25 reusable constraint components demonstrates a scalable approach to validating complex domain ontologies, which we expect to contribute for facilitate adoption of RDF intents by the intent-based automation community in telecom networks.

\paragraph{Resource Availability.} \tioshacl{} is available at \url{https://github.com/EricssonResearch/tio-shacl} under the \MIT{} license. The repository includes all \shacl{} shapes, the reusable constraint library, 133 test cases, and validation tooling.